\documentclass{article}

    \PassOptionsToPackage{numbers, compress}{natbib}
 \usepackage[preprint]{neurips_2026}

\usepackage[utf8]{inputenc} 
\usepackage[T1]{fontenc}    
\usepackage{hyperref}       
\usepackage{url}            
\usepackage{booktabs}       
\usepackage{amsfonts}       
\usepackage{nicefrac}       
\usepackage{microtype}      
\usepackage[dvipsnames]{xcolor}         

\usepackage{amsmath}
\usepackage{amssymb}
\usepackage{mathtools}
\usepackage{amsthm}
\usepackage{enumitem}
\usepackage{graphicx}
\usepackage{multirow}
\usepackage{subcaption}
\usepackage{wrapfig}
\usepackage{pifont}
\usepackage{adjustbox}

\newif\ifrevcolors
\revcolorstrue          

\definecolor{revAColor}{RGB}{30,90,200}    
\definecolor{revBColor}{RGB}{130,30,160}   
\definecolor{revCColor}{RGB}{210,90,10}    
\definecolor{revDColor}{RGB}{20,120,40}    
\definecolor{revTColor}{RGB}{180,30,90}    

\ifrevcolors
  \newcommand{\rA}[1]{{\color{black}#1}}   
  \newcommand{\rB}[1]{{\color{black}#1}}   
  \newcommand{\rC}[1]{{\color{black}#1}}   
  \newcommand{\rD}[1]{{\color{black}#1}}   
  \newcommand{\rT}[1]{{\color{black}#1}}   
  \newcommand{\rAS}[1]{{\color{black}#1}}   
\else
  \newcommand{\rA}[1]{#1}
  \newcommand{\rB}[1]{#1}
  \newcommand{\rC}[1]{#1}
  \newcommand{\rD}[1]{#1}
  \newcommand{\rT}[1]{#1}
  \newcommand{\rAS}[1]{#1}
\fi

\title{GLASS: Global-Local Aggregation for \\ Inference-time Sparsification of LLMs}

\author{%
  Amirmohsen Sattarifard\thanks{Equal contribution} \quad Sepehr Lavasani\textsuperscript{*} \quad Kunlin Zhang \quad Amirhossein Rajabpour\\
  Huawei Technologies Canada Co., Ltd.
  \AND
  Hanlin Xu  \quad Fengyu Sun\\
  Huawei Technologies Ltd.
  \And
  Negar Hassanpour\thanks{Equal advising}  \quad Chao Gao\textsuperscript{$\dagger$}\\
  Huawei Technologies Canada Co., Ltd.
}

\begin{document}

\maketitle

\begin{abstract}
Inference-time sparsification is a promising path to deploy large language models~(LLMs) on resource-constrained devices, yet existing training-free methods typically estimate feedforward network~(FFN) neuron importance from the input prompt alone. We show this prompt-only signal is often unreliable, especially for short prompts and long-form decoding, leading to inaccurate masks and degraded generation fidelity. We propose GLASS, a plug-and-play, training-free framework that stabilizes dynamic FFN pruning by aggregating two complementary views of neuron criticality: local prompt-specific activations and a global model-intrinsic prior. GLASS fuses global and local signals via rank aggregation, yielding robust critical-neuron selection even when the prompt is short. We interpret GLASS as the maximum-a-posteriori consensus ranking under a permutation-based probabilistic model, providing a principled foundation for its weighted rank-aggregation rule. We apply GLASS to a diverse set of open-source LLMs, and show that it yields substantial improvements over prior training-free baselines in the challenging short-prompt, long-generation scenarios, achieving up to 45.10\% lower perplexity and 25.73\% lower KL divergence, while delivering significant on-device decoding speedup.
\end{abstract}

\section{Introduction}
\label{sec:intro}

Large Language Models~(LLMs) have achieved strong performance across a wide range of language tasks~\citep{team2023gemini,team2024gemma,touvron2023llama,jiang2023mistral7b,anthropic2024claude}, motivating their deployment on diverse consumer and enterprise devices~\citep{alizadeh2024llm,song2024powerinfer,xue2024powerinfer}. Yet, their large parameter counts translate into substantial memory traffic and compute demand, making low-latency inference on consumer-grade hardware difficult~\citep{xue2024powerinfer}. In practice, some form of pruning is often required to reduce latency and memory footprint while preserving quality.

Depending on how weights are removed or skipped, pruning methods are commonly categorized as \emph{static} or \emph{dynamic}. Static pruning~\citep{lecun1989optimal,ma2023llm,xia2023sheared} permanently deletes a subset of parameters (one-shot or iteratively), typically followed by training/fine-tuning to recover performance. After deployment, the sparsity pattern is fixed and does not depend on the input. In contrast, dynamic pruning~\citep{song2024powerinfer,xue2024powerinfer,alizadeh2024llm} conditionally skips computation based on the current input, so different requests may activate different subsets of the model.

A straightforward dynamic approach is to train a \emph{predictor} that estimates layer-wise activation patterns and prunes computation accordingly. While effective, predictor-based methods incur (i) an extra training stage, (ii) inference-time overhead, and (iii) potentially higher memory-transfer overhead due to frequent (un)loading of neurons.
As a result, \emph{training-free} dynamic pruning (i.e., without training an auxiliary predictor and without modifying the LLM weights) is often preferred in deployment settings.

\begin{wrapfigure}{r}{0.51\textwidth}
\vspace{-0.45em}
  \centering
    \includegraphics[width=0.51\columnwidth]{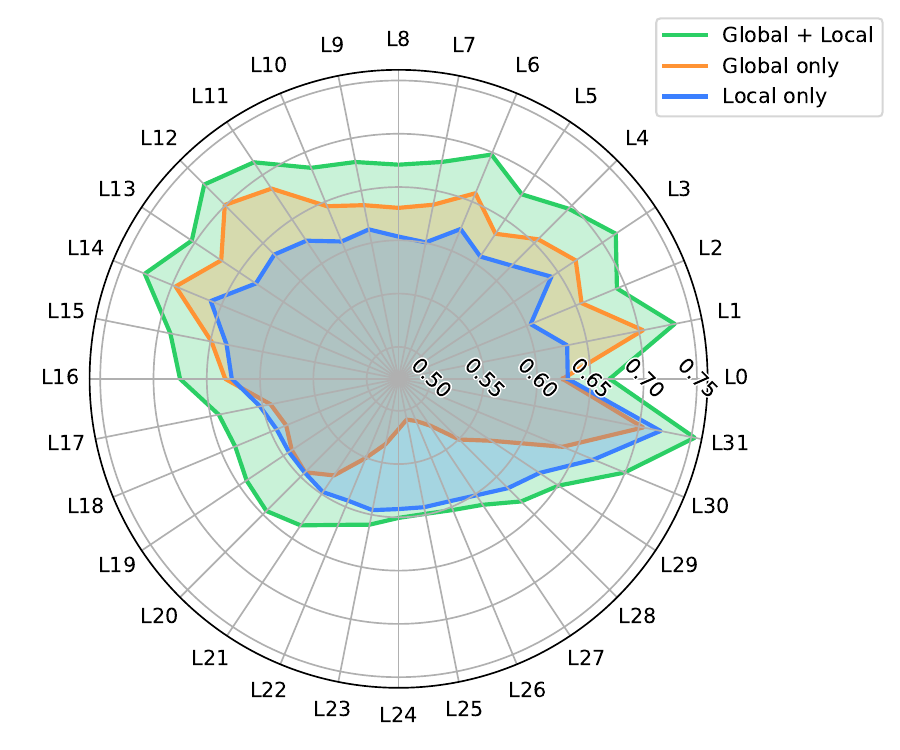}
    \caption{
    Jaccard similarity~($\uparrow$) between critical neuron sets identified by the Global-Local Aggregation~(green), Global~(orange), and Local~(blue) methods using the oracle set as reference, across various layers on \hbox{Llama 3 8B}.
    Aggregating global and local information allows GLASS to more faithfully identify the oracle critical neuron set 
    (see Sec.~\ref{sec:how_ga} and App.~\ref{app:how_ga} for details).
    }
    \label{fig:jacc_radar}
\end{wrapfigure}


In LLMs, training-free FFN pruning methods~\citep{dong2024prompt,ma2024first} estimate a prompt-derived sparsity mask and reuse it during decoding, relying on the \emph{flocking phenomenon}~\citep{ma2024first}. However, with short prompts or long responses, activation patterns may drift from the prefill mask. Updating masks online~\citep{zhang2025r,you2025spark,yin2025duogpt,storai2025smarter,qi2025deltallm,shin2025sparseinfer,lee2025recap,yang2025sparse} can mitigate this drift, but reduces the aggressively pruned portion of decoding and limits end-to-end speedups~\citep{liu2024teal}. \rC{It also sacrifices a key edge-deployment benefit of static masks: keeping a fixed compact FFN subset resident in fast memory and avoiding repeated I/O.}

\begin{wrapfigure}{r}{0.55\textwidth}
\vspace{-1.5em}
  \centering
    \includegraphics[width=0.45\linewidth]{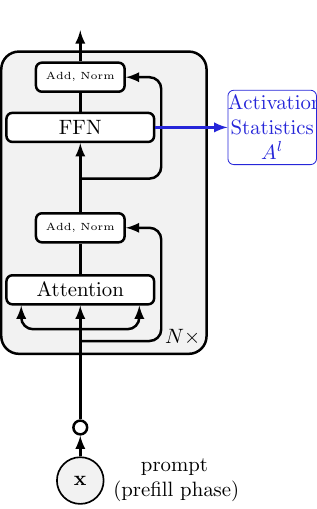} \quad
    \includegraphics[width=0.4925\linewidth]{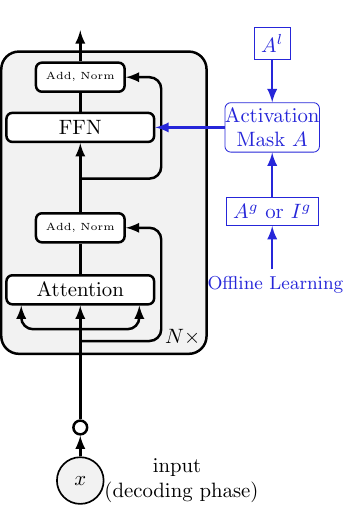}
    \caption{Overview of GLASS. We (left) learn a robust mask by fusing local (prompt) evidence with a global prior from NPS, and (right) apply the resulting mask for FFN activation pruning during generation.}
    \label{fig:glass_illustrate}
\end{wrapfigure}

We address this reliability-efficiency tension with \textbf{GLASS}, a simple global-local FFN sparsification framework. 
GLASS augments prompt-derived neuron importance with an independent \emph{global} prior, computed once offline via NPS without task-specific data or model training.
Using Null-Prompt Stimulation (NPS; see Sec.~\ref{sec:nps}), GLASS estimates either neuron activation magnitude or performance impact, yielding \textbf{\hbox{A-GLASS}} and \textbf{\hbox{I-GLASS}}, respectively. At inference, it fuses this global prior with local prompt-specific activations to produce masks robust to short prompts and long generations. An overview is shown in Fig.~\ref{fig:glass_illustrate}, and Fig.~\ref{fig:jacc_radar} shows that global-local aggregation improves alignment with the oracle active-neuron mask.

\paragraph{Contributions.}
We make \rC{four} contributions:
(i)~we \rB{identify a previously unrecognized failure mode of prompt-only training-free FFN sparsification: short-prompt masks drift from decoding-time activations, hurting long-form generation;}
(ii)~we introduce \textbf{GLASS}, a plug-and-play, training-free inference-time pruning framework that fuses \emph{local} prompt evidence with a \emph{global} model-intrinsic prior, yielding reliable FFN masks under short prompts and long generations;
(iii)~\rB{we provide a probabilistic consensus-ranking interpretation of GLASS, showing that under a Mallows-type model with squared Spearman rank distance, its weighted rank aggregation is the MAP estimator;} and
(iv)~we achieve up to \textbf{45.10\%} lower perplexity and \textbf{25.73\%} lower KL divergence than the strongest training-free baseline at 50\% sparsity, plus substantial on-device decoding phase speedup \rD{(up to $\sim\!11\times$ on Samsung Galaxy S25 Ultra when memory residency dominates)}.%

\section{Assumptions and Objectives} \label{sec:problem}
\subsection{Model Structure Assumption} \label{sec:preliminary}
We consider a modern Transformer~\citep{NIPS2017_3f5ee243} model composed by stacking of multiple layers each consisting of an attention block and a FFN block. Each FFN block adopts a gated structure:
\begin{equation}
    \begin{aligned}
        \mathbf{z}_u &= \mathbf{x}W_{\text{up}} + \mathbf{b}_{\text{up}}, &
        \mathbf{a}_u &= \phi_u(\mathbf{z}_u) \\
        \mathbf{z}_g &= \mathbf{x}W_{\text{gate}} + \mathbf{b}_{\text{gate}}, &
        \mathbf{a}_g &= \phi_g(\mathbf{z}_g) \\
        \mathbf{h} &= \mathbf{a}_u \odot \mathbf{a}_g, &
        \mathbf{y} &= \mathbf{h}W_{\text{down}} + \mathbf{b}_{\text{down}},
    \end{aligned}
    \label{eq:gated_ffn}
\end{equation}

Here, \(\mathbf{x} \in \mathbb{R}^{d}\) is the input token embedding. The FFN expands the representation to a higher-dimensional space of width \(m\), performs elementwise gating, and projects back to \(\mathbb{R}^{d}\):
\begin{itemize}[nosep]
    \item \(W_{\text{up}}, W_{\text{gate}} \in \mathbb{R}^{d \times m}\) are the expansion matrices.
    \item \(W_{\text{down}} \in \mathbb{R}^{m \times d}\) is the projection matrix.
    \item \(\mathbf{b}_{\text{up}}, \mathbf{b}_{\text{gate}} \in \mathbb{R}^m\), and \(\mathbf{b}_{\text{down}} \in \mathbb{R}^d\) are the bias terms.
    \item \(\phi_u(\cdot)\) is an activation function and \(\phi_g(\cdot)\) is a sigmoid-like gating function.
    \item \(\mathbf{h} \in \mathbb{R}^m\) is the FFN hidden unit vector. Its \textit{j}-th element is denoted by \(h_j\)
\end{itemize}

\textbf{Associated weights of a hidden unit.}
Each hidden unit \(h_j\), \(j \in \{1,\dots,m\}\), is associated with the \(j\)-th up-projection column ($W_{\mathrm{up}}^{(:,j)}$), 
the optional \(j\)-th gate-projection column ($W_{\mathrm{gate}}^{(:,j)}$), 
and the \(j\)-th down-projection row ($W_{\mathrm{down}}^{(j,:)}$).
For each output dimension \(k \in \{1,\dots,d\}\), the FFN output can be written as
\begin{equation}
    y_k = \sum_{j=1}^{m} \phi_u\!\left(\mathbf{x} W_{\mathrm{up}}^{(:,j)} + b_{\mathrm{up}}^{(j)}\right) \cdot g_j \cdot W_{\mathrm{down}}^{(j,k)} + b_{\mathrm{down}}^{(k)}, \quad k \in \{1,\dots,d\},
    \label{eq:ffn_output_dim_k}
\end{equation}
where
\begin{equation}
    g_j = \begin{cases} \phi_g\!\left(\mathbf{x} W_{\mathrm{gate}}^{(:,j)} + b_{\mathrm{gate}}^{(j)}\right), & \text{if a gate branch is used}, \\ 1, & \text{otherwise}. \end{cases}
\end{equation}

Eq.~\eqref{eq:ffn_output_dim_k} expresses the FFN output as an additive decomposition over the hidden dimension \(j\). 
Sparsifying the FFN corresponds to removing or masking a subset of hidden units \(j\), thereby eliminating their corresponding summands in the FFN output.

\subsection{Dynamic Activation for FFN}
The objective is to dynamically prune a subset of FFN hidden units (neurons) at inference-time, while minimizing degradation in model quality with respect to the base (unpruned) model. For each FFN block, given a budget \(k < m\), we aim to select a subset of \(k\) hidden units (referred to as \emph{critical neurons}). This can be represented as a binary mask with $k$ non-zero entries. In the decoding phase, with the mask, model inference becomes faster due to reduced computation for each FFN. 

\textbf{Computation and Storage Overhead.}
\textcolor{black}{The computation and storage overhead of dynamic FFN activation is small, requiring only a 1D binary mask of size $m$ for each FFN layer. This overhead is almost negligible in comparison to the whole LLM.} 
\section{GLASS: Global-Local Aggregation for FFN Sparsification} 
\label{sec:gni}

Although it is sensible to add a global measurement for assisting neuron pruning, how to compute such a metric is still unclear. A suitable definition of global importance (noted as $M^{g}$) is also unclear. In GLASS, we propose two definitions for model-intrinsic neuron importance. The first one is defined by measuring activation magnitude (noted as $A^{g}$), and the second is by measuring performance impact ($I^{g}$).  For both definitions, given a model $\mathcal{M}$, we propose a null prompt stimulation (NPS) algorithm to reliably compute each metric. NPS not only eliminates the need for offline datasets, but also produces more accurate model-intrinsic importance estimates.  After introducing $A^{g}$, $I^{g}$, and NPS, we describe in detail how GLASS aggregates them with local importance learned during the prefill phase.

\subsection{Activation Magnitude $A^{g}$ as Global Importance} \label{sec:ga}
Similar to learning importance scores at prefill, we can use average activation on a global stimulation set as global importance.  For each FFN unit $j$, we can thus define its \emph{global activation magnitude} as:
\begin{equation}
    A^{g}_j := \mathbb{E}_{\mathbf{x} \sim \mathcal{D}} \left[ | \hat{h}_j(\mathbf{x}) | \right],
    \label{eq:global-activation}
\end{equation}
where \(\hat{\mathbf{h}}(\mathbf{x})=\mathbf{h}(\mathbf{x})/(\|\mathbf{h}(\mathbf{x})\|_2+\epsilon)\) denotes the \(\ell_2\)-normalized activation vector for token \(\mathbf{x}\).
The $\ell_2$ normalization is to ensure that activations are comparable across tokens and layers.
The resulting $A^{g}_j$ serves as a model-intrinsic importance score for unit $j$, 
and will be used for critical neuron selection. 

\subsection{Impact $I^{g}$ as Global Importance}\label{sec:gi}
Alternatively, we propose to define the global importance of a neuron by considering how it impacts a given loss, echoing classical Taylor pruning~(e.g.,~\citep{figurnov2016perforatedcnns,molchanov2016pruning}). Let \(f_\theta(\textbf{h}(\mathbf{x}))\) be the network output, 
\(\mathbf{y}\) the target token (label),
and \(\mathcal{L} \bigl(f_\theta(\textbf{h}(\mathbf{x})),\mathbf{y}\bigr)\) the cross-entropy loss of the model as a function of the activations
of the current FFN layer \(\mathbf{h}(x)=(h_1(x),\dots,h_m(x))\).
Ablating neuron \(j\) sets \(h_j\mapsto0\), i.e. 
\(\Delta h_j=-h_j\).
Applying a first-order Taylor expansion of \(\mathcal{L}\) at \(\mathbf{h}\) yields
\begin{equation}
\Delta\mathcal{L}_j
\;\approx\;
\nabla_{\mathbf{h}}\mathcal{L}(\mathbf{h})\;\cdot\;\Delta\mathbf{h}
\;=\;
\frac{\partial\mathcal{L}}{\partial h_j}\,\Delta h_j
\;=\;
-\,h_j\,\delta_j,
\label{eq:global-impact-loss}
\end{equation}
where \(\delta_j \coloneqq \partial\mathcal{L}/\partial h_j\).
Because only the magnitude of $\Delta\mathcal{L}_j$ in Eq.~\eqref{eq:global-impact-loss} matters for pruning, we define the
(neuron-wise) \emph{impact score}
\begin{equation}
I_j^g
\;=\;
\mathbb{E}_{\mathbf{x}\sim\mathcal{D}}
\bigl[\,
\lvert h_j(\mathbf{x})\,\delta_j(\mathbf{x})\rvert
\bigr].
\label{eq:global-impact}
\end{equation}

\rAS{Neurons with larger $I^g_j$ are expected to incur a larger local disturbance in loss when removed. As in standard Taylor pruning, this score is a first-order saliency proxy. Since GLASS removes multiple neurons simultaneously, $I^g_j$ should be interpreted as a ranking heuristic rather than an exact estimate of the full-mask loss change. The resulting $I^g_j$ serves as another model-intrinsic importance score and will be used for critical neuron selection.}

\subsection{Null Prompt Stimulation to Compute $I^{g}$ and $A^{g}$} \label{sec:nps}
To compute global activations or neural impacts, one approach is to run the model on a dataset and collect the respective statistics~\citep{lee2404cats}. 
However, standard corpora like Wikipedia can bias the statistics toward specific linguistic patterns.
We instead propose Null Prompt Stimulation~(NPS): 
generating text from the model itself using a null prompt (e.g., empty string or BOS token) under typical sampling settings (implementation details in App.~\ref{app:imp-det}). 
\rB{Importantly, NPS is not merely ``using no dataset''; by sampling from the model's own predictive distribution given minimal conditioning, it directly probes the model's intrinsic neuron utilization, eliminating any external corpus bias. Sec.~\ref{sec:nps-helps} shows that NPS-derived priors consistently outperform WikiText-derived priors at every density level (see Tab.~\ref{tab:kld_combined}). } \rAS{For I-GLASS, gradients \(\delta_j\) are computed by teacher-forcing the generated NPS sequence and using each self-generated next token as the pseudo-label; full details are in App.~\ref{app:imp-det}.}


\subsection{Fusion of Global and Local Importance}
\label{sec:glass:formulation}

\begin{wrapfigure}{r}{0.55\textwidth}
\vspace{-2.5em} 
  \centering
  \includegraphics[width=0.55\columnwidth]{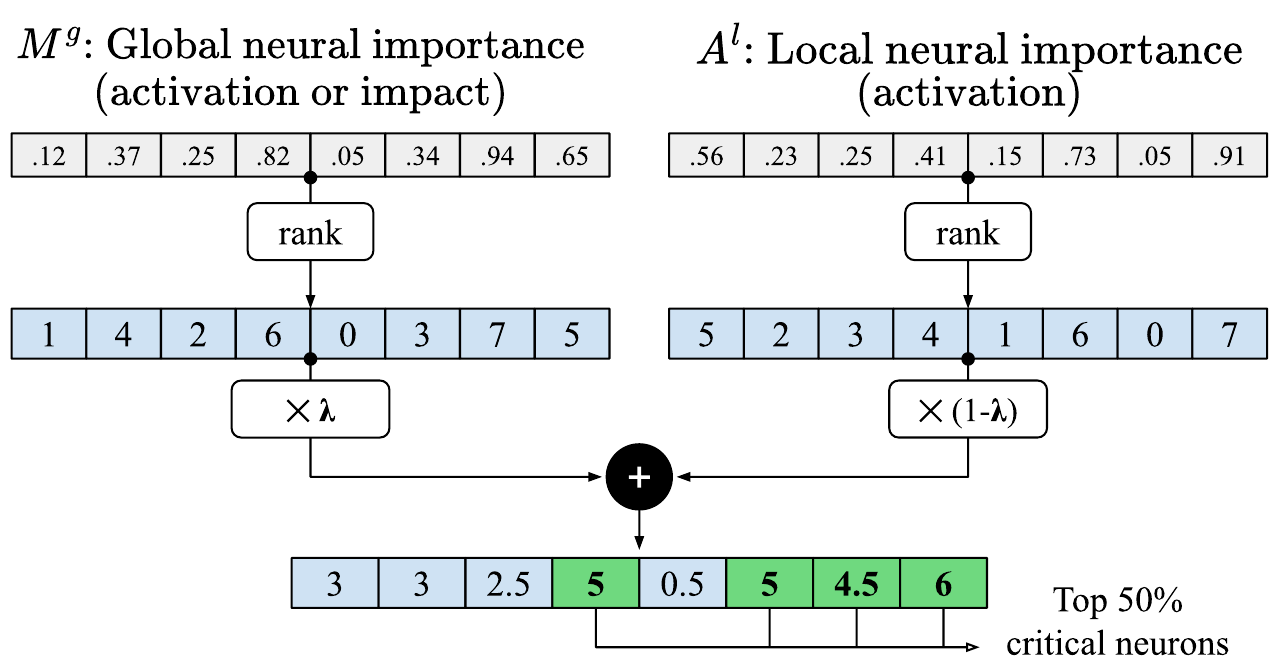}
  \caption{Rank aggregation in GLASS.}
  \label{fig:bigfig}
\end{wrapfigure}

Assume each FFN layer contains $m$ neurons. For each neuron $j \in [m]$, let $A^{l}_j$ denote the \emph{expected activation magnitude} of FFN unit $j$ over the prompt tokens, and let $M^{g}_j$ denote its \emph{global importance} obtained either from activation magnitude $A^g$ or impact $I^g$ through NPS: $A^{l}_j, M^{g}_j \in \mathbb{R}_{\ge0}$. Directly fusing these heterogeneous quantities can be problematic because their scales, distributions, and semantics differ. \rC{This is particularly acute when fusing an activation-based local signal with an impact-based global signal, which live on an entirely different scales.} We therefore first convert both signals into rank space. 

Specifically, let $\operatorname{rank}_{\uparrow}$ assign rank $1$ to the smallest value and rank $m$ to the largest value, and define $R^{(l)}_j = \operatorname{rank}_{\uparrow}(A^{l})_j$, $R^{(g)}_j = \operatorname{rank}_{\uparrow}(M^{g})_j.$
Thus, smaller importance scores receive smaller ranks, and larger $R^{(l)}_j$ or $R^{(g)}_j$ indicates higher importance, with rank $m$ corresponding to the most important neuron under each signal.%
\footnote{
\rAS{If exact ties occur before rank conversion, we use stable deterministic tie-breaking by neuron index to obtain a valid permutation. In practice, exact ties are rare because $A^l$, $A^g$, and $I^g$ are floating-point averages over many tokens. If a tie occurs at the top-$k$ boundary, the same deterministic rule is used to ensure reproducible neuron selection.}
}
Ranking makes the local and global signals comparable while preserving their ordinal preferences, and makes the fusion invariant to monotone transformations of either importance measure.

\rC{We now give a probabilistic interpretation of GLASS as a maximum-a-posteriori (MAP) consensus ranking under a permutation-based model.} Let $\pi$ denote the latent consensus permutation over neurons in a layer, and let $\pi^{(l)}$ and $\pi^{(g)}$ denote the observed local and global permutations induced by sorting the local and global importance scores, respectively. Here, $\pi$ is an ordered list of neurons from least important to most important. Let $r(\pi)\in\{1,\dots,m\}^m$ denote the corresponding rank vector, where the $j$-th entry gives the rank position assigned to neuron $j$; rank $1$ corresponds to the least important neuron and rank $m$ corresponds to the most important neuron. Thus, larger rank values indicate higher importance.

Following Mallows-type distance-based ranking models~\citep{mallows1957non,fligner1986distance}, we model the two observed permutations as conditionally independent noisy observations of $\pi$ using squared Spearman rank distance~\citep{diaconis1977spearman}:
\begin{equation*}
P(\pi^{(l)} \mid \pi)
\propto
\exp\!\left(
-\beta_l \, \|r(\pi^{(l)}) - r(\pi)\|_2^2
\right),
\qquad
P(\pi^{(g)} \mid \pi)
\propto
\exp\!\left(
-\beta_g \, \|r(\pi^{(g)}) - r(\pi)\|_2^2
\right).
\label{eq:mallows-spearman}
\end{equation*}
The MAP%
\footnote{
\rAS{For this probabilistic interpretation, we assume a uniform prior over the latent consensus permutation $\pi$, so MAP estimation reduces to maximizing the likelihood.}
}
consensus permutation is therefore
\[
\pi^\star
=
\arg\min_{\pi}
\beta_l \|r(\pi^{(l)}) - r(\pi)\|_2^2
+
\beta_g \|r(\pi^{(g)}) - r(\pi)\|_2^2.
\]
\rAS{Note that we treat the local and global rankings as conditionally independent given $\pi$. This conditional-independence assumption is a simplifying modeling approximation: both rankings are derived from the same model and need not be statistically independent in a strict sense. Its role is to yield a tractable additive consensus objective. Empirically, our Local-Only, Global-Only, and Global-Local ablations in Sec.~\ref{sec:how_ga} support that the two rankings contain complementary information rather than one signal simply dominating the other.} 

As shown in App.~\ref{app:glass-consensus-proof}, because every feasible rank vector is a permutation of $(1,\dots,m)$, the squared-norm $\|r(\pi)\|_2^2$ is constant across all candidate $\pi$, so the MAP solution reduces (via the rearrangement inequality) to assigning larger consensus ranks to larger values of $s_j = \beta_l R_j^{(l)} + \beta_g R_j^{(g)}.$
Equivalently, the neurons are ordered from most important to least important by sorting $s_j$ in descending order. Normalizing by the positive constant $\beta_l+\beta_g$ does not change the induced ranking. With $\lambda = \frac{\beta_g}{\beta_l+\beta_g}$,
we obtain the GLASS score
\begin{equation}
\mathrm{GLASS}_j
=
(1-\lambda) R_j^{(l)} + \lambda R_j^{(g)}.
\label{eq:ham}
\end{equation}

\rAS{This is a weighted Borda-style rank aggregation rule~\citep{borda1781memoire}: each neuron receives a weighted sum of its local and global ranks, and larger $\mathrm{GLASS}_j$ indicates higher consensus importance.} In practice, GLASS does not require the full permutation; it retains the $k$ neurons with the largest scores under this induced consensus ordering.


When $M^{g}_j$ represents global activation magnitude (Eq.~\eqref{eq:global-activation}), we refer to the resulting method as \hbox{A-GLASS}; when it represents neural impact (Eq.~\eqref{eq:global-impact}), we refer to it as \hbox{I-GLASS}. 
\rAS{We set $\lambda = 0.5$ by default, corresponding to equal reliability of local and global evidence. This balances input-specific but short-prompt-sensitive local ranks with stable but input-agnostic global ranks, preserving the plug-and-play, training-free nature of GLASS. 
Our sensitivity analysis in App.~\ref{app:lambda-sweep} shows that performance varies smoothly with $\lambda$ and remains strong near $\lambda=0.5$ on representative models.} 

\section{Experiments} \label{sec:experiments}

Following prior work, we evaluate GLASS on both Classification~(C) and Short-form Generation~(SG) benchmarks. To fully show the merits of GLASS, we also introduce and evaluate on a Long-form Generation~(LG) benchmark consisting of brief prompts and lengthy outputs. This makes LG ideal for comparing sparsification quality in long-generation scenarios.

\paragraph{Evaluation Datasets.}
C datasets include \textbf{HellaSwag}~\citep{zellers2019hellaswag}, \textbf{PIQA}~\citep{Bisk2020}, \textbf{COPA}~\citep{roemmele2011choice}, \textbf{ARC-E/C}~\citep{allenai:arc}, and \textbf{BoolQ}~\citep{clark2019boolq}. 
For SG, we use \textbf{XSum}~\citep{Narayan2018DontGM}, \textbf{CNN/DailyMail}~\citep{see-etal-2017-get,DBLP:conf/nips/HermannKGEKSB15}, \textbf{CoQA}~\citep{reddy-etal-2019-coqa}, and \textbf{QASPER}~\citep{dasigi2021dataset}. 
Details of the C and SG datasets are provided in the App.~\ref{app:benchmark-details}.

\begin{table*}[t] 
\centering
\caption{Classification tasks (0-shot unnormalized accuracy) and short-generation at $50\%$ FF sparsity.} \label{tab:sg}
\scalebox{0.6}{
\begin{tabular}{@{}lccccccc|cccc@{}}
\toprule
& & \multicolumn{6}{c}{Classification} & \multicolumn{4}{c}{Short-Generation} \\

\multicolumn{1}{c}{}            &         & \multirow{2}{*}{HellaSwag} & \multirow{2}{*}{PIQA} & \multirow{2}{*}{COPA} & \multirow{2}{*}{ARC-e} & \multirow{2}{*}{ARC-c} & \multirow{2}{*}{BoolQ}  & XSum                   & CNN/DailyMail          & CoQA          & QASPER \\                                                                                                                                                                                                                                                                                                                                                                                 &         &                               &                          &                          &                           &                          &                             & Rouge-1 / 2 / L        & Rouge-1 / 2 / L        & F1 / EM       & F1     \\                                                                                                                                                                                                                                                                                                                                                 \toprule                                                                                                                                                                                                                                                                                                                                                                                                                                                                                                                                                                                                                                           \multirow{2}{*}{Gemma 7B}       & I-GLASS & 60.59                         & 80.25                    & 88.00                    & 81.69                     & 49.83                     & 81.56            & 25.74 / 7.69 /   20.75 & 18.97 / 4.78 /   17.31 & 78.20 / 64.67 & 67.13         \\
                                & GRIFFIN & 60.62                         & 80.03                    & 88.00                    & 81.99                     & 50.09                     & 81.56            & 25.36 / 7.67 /   20.48 & 18.26 / 4.74 /   16.75 & 78.27 / 64.05 & 67.13         \\
\midrule
                                                                                                                                                                                                                 \multirow{2}{*}{Llama2 7B}      & I-GLASS & 57.14                         & 78.07                    & 86.00                    & 76.30                     & 43.52                     & 77.74            & 27.18 / 9.06 /   22.62 & 10.07 / 0.13 /   9.55  & 77.35 / 63.88 & 67.31         \\                                                                                                                                                                   & GRIFFIN & 57.14                         & 78.07                    & 86.00                    & 76.30                     & 43.52                     & 77.74            & 27.18 / 9.06 /   22.62 & 10.07 / 0.13 /   9.55  & 77.35 / 63.88 & 67.31         \\
\midrule                                                                                                                                                                                                         
\multirow{2}{*}{Mistral 7B}     & I-GLASS & 61.01                         & 80.47                    & 92.00                    & 79.76                     & 50.77                     & 80.18            & 27.27 / 8.70 /   22.37 & 0.59 / 0.06 /   0.55   & 80.53 / 67.70 & 72.73         \\
                                & GRIFFIN & 60.97                         & 80.69                    & 92.00                    & 79.71                     & 50.09                     & 80.37            & 27.03 / 8.95 /   22.39 & 1.03 / 0.20 /   0.92   & 80.39 / 66.93 & 72.73         \\                                                                                                                                   \midrule                                                                                                                                                                                                                                                                                                                                                                                                                          \multirow{2}{*}{OPT 6.7B}       & I-GLASS & 50.45                         & 75.90                    & 80.00                    & 63.93                     & 30.72                     & 65.93            & 19.95 / 4.54 /   16.47 & 12.27 / 0.69 /   11.59 & 68.42 / 54.85 & 64.54         \\
                                & GRIFFIN & 50.49                         & 75.73                    & 80.00                    & 63.89                     & 30.63                     & 65.41            & 21.18 / 5.36 /   17.68 & 12.99 / 1.05 /   12.25 & 69.03 / 55.00 & 64.54         \\
\midrule                                                                                                                                                                                                         
\multirow{2}{*}{ReLU-Llama2 7B} & I-GLASS & 53.95                         & 77.42                    & 88.00                    & 74.62                     & 39.25                     & 77.98            & 24.66 / 7.45 /   20.36 & 21.15 / 6.83 /   19.48 & 78.49 / 66.73 & 60.35         \\
                                & GRIFFIN & 53.95                         & 77.42                    & 88.00                    & 74.62                     & 39.25                     & 77.98            & 24.66 / 7.45 /   20.36 & 21.15 / 6.83 /   19.48 & 78.49 / 66.73 & 60.35         \\
\bottomrule
\end{tabular}
}
\label{tab:g_cls}
\vspace{2.5em}
\centering
\caption{
    Perplexity~(PPL) and top-100 KL Divergence~(KLD) for GRIFFIN compared to our \hbox{A/I-GLASS}~(NPS).
    The Imp\% columns represent the improvement percentage over GRIFFIN.
}
\scalebox{0.8}{
\begin{tabular}{@{}llrrrrrrrrr@{}}
\toprule
\textbf{Model}                               & \textbf{Metric} & \multicolumn{1}{c}{\textbf{GRIFFIN}} & \multicolumn{1}{c}{\textbf{A-GLASS}} & \multicolumn{1}{r}{\textbf{$\rightarrow$ Imp\%}} & \multicolumn{1}{c}{\textbf{I-GLASS}} & \multicolumn{1}{r}{\textbf{$\rightarrow$ Imp\%}} \\ \midrule
\multirow{2}{*}{\textbf{Gemma 7B}}           & PPL             & 3.7014 \small{(0.0316)}                      & 3.5635 \small{(0.0368)}                      & 3.73\%                             & 3.3982 \small{(0.0310)}                      & \textbf{8.19\%}                    \\
                                             & KLD             & 0.6453 \small{(0.0033)}                      & 0.5998 \small{(0.0037)}                      & 7.05\%                             & 0.5661 \small{(0.0036)}                      & \textbf{12.27\%}                   \\ \midrule
\multirow{2}{*}{\textbf{Gemma 2 9B}}         & PPL             & 4.0143 \small{(0.0262)}                      & 3.5105 \small{(0.0217)}                      & 12.55\%                            & 3.4962 \small{(0.0218)}                      & \textbf{12.91\%}                   \\
                                             & KLD             & 0.6970 \small{(0.0026)}                      & 0.5877 \small{(0.0025)}                      & 15.68\%                            & 0.5821 \small{(0.0026)}                      & \textbf{16.48\%}                   \\ \midrule
\multirow{2}{*}{\textbf{Gemma 2 27B}}        & PPL             & 2.7729 \small{(0.0190)}                      & 2.4860 \small{(0.0165)}                      & 10.35\%                            & 2.4268 \small{(0.0155)}                      & \textbf{12.48\%}                   \\
                                             & KLD             & 0.4338 \small{(0.0024)}                      & 0.3515 \small{(0.0025)}                      & 18.97\%                            & 0.3349 \small{(0.0024)}                      & \textbf{22.80\%}                   \\ \midrule
\multirow{2}{*}{\textbf{Gemma 3n E2B}}       & PPL             & 23.1611 \small{(0.2938)}                     & 12.7155 \small{(0.1806)}                     & \textbf{45.10\%}                   & 14.4240 \small{(0.2159)}                     & 37.72\%                            \\
                                             & KLD             & 1.9983 \small{(0.0063)}                      & 1.4841 \small{(0.0068)}                      & \textbf{25.73\%}                   & 1.5677 \small{(0.0072)}                      & 21.55\%                            \\ \midrule
\multirow{2}{*}{\textbf{Gemma 3n E4B}}       & PPL             & 8.5984 \small{(0.0920)}                      & 5.9075 \small{(0.0707)}                      & \textbf{31.30\%}                   & 6.3256 \small{(0.0809)}                      & 26.43\%                            \\
                                             & KLD             & 1.3106 \small{(0.0053)}                      & 0.9796 \small{(0.0061)}                      & \textbf{25.26\%}                   & 1.0190 \small{(0.0066)}                      & 22.25\%                            \\ \midrule
\multirow{2}{*}{\textbf{Llama 3 8B}}         & PPL             & 5.2523 \small{(0.0434)}                      & 4.0374 \small{(0.0324)}                      & 23.13\%                            & 4.0295 \small{(0.0331)}                      & \textbf{23.28\%}                   \\
                                             & KLD             & 0.9426 \small{(0.0047)}                      & 0.7407 \small{(0.0044)}                      & 21.42\%                            & 0.7354 \small{(0.0046)}                      & \textbf{21.98\%}                   \\ \midrule
\multirow{2}{*}{\textbf{Mistral 7B}}         & PPL             & 5.0059 \small{(0.0407)}                      & 4.4786 \small{(0.0384)}                      & \textbf{10.53\%}                   & 4.4860 \small{(0.0387)}                      & 10.39\%                            \\
                                             & KLD             & 0.8774 \small{(0.0041)}                      & 0.7615 \small{(0.0045)}                      & 13.21\%                            & 0.7593 \small{(0.0046)}                      & \textbf{13.46\%}                   \\ \midrule
\multirow{2}{*}{\textbf{Qwen 2.5 7B}}        & PPL             & 5.0840 \small{(0.0378)}                      & 5.5823 \small{(0.0581)}                      & -9.80\%                            & 5.5424 \small{(0.0593)}                      & -9.02\%                            \\
                                             & KLD             & 0.6772 \small{(0.0028)}                      & 0.6880 \small{(0.0039)}                      & -1.59\%                            & 0.6684 \small{(0.0040)}                      & \textbf{1.30\%}                    \\ \midrule
\multirow{2}{*}{\textbf{Qwen 2.5 14B}}       & PPL             & 5.4330 \small{(0.0882)}                      & 5.5823 \small{(0.0677)}                      & -2.75\%                            & 5.0279 \small{(0.0701)}                      & \textbf{7.46\%}                    \\
                                             & KLD             & 0.6557 \small{(0.0048)}                      & 0.5947 \small{(0.0048)}                      & 9.30\%                             & 0.5680 \small{(0.0050)}                      & \textbf{13.38\%}                   \\ \midrule
\multirow{2}{*}{\textbf{Phi 3 14B (medium)}} & PPL             & 3.8981 \small{(0.0328)}                      & 4.1384 \small{(0.0283)}                      & -6.16\%                            & 3.7364 \small{(0.0237)}                      & \textbf{4.15\%}                    \\
                                             & KLD             & 0.6763 \small{(0.0034)}                      & 0.6928 \small{(0.0035)}                      & -2.44\%                            & 0.6188 \small{(0.0031)}                      & \textbf{8.50\%}                    \\ \bottomrule
\end{tabular}
}
\label{tab:main}
\end{table*}

For the LG benchmark, we use \textbf{Alpaca}~\citep{alpaca}, a self-instruction dataset generated with \hbox{GPT-3}. \rD{Alpaca is suitable because it spans diverse tasks, pairs short prompts with long outputs, and is widely used for reproducible LLM evaluation.} To focus on long generation, we keep samples with ground-truth outputs over 100 tokens and exclude cases where any studied base model generates fewer than 20 tokens, yielding 3,602 samples.


\paragraph{Evaluation Models.}
For C and SG tasks, we evaluate five publicly available LLMs in the 6-7 billion parameter range:
\textbf{Gemma~7B}~\citep{team2024gemma},
\textbf{Llama 2~7B}~\citep{touvron2023llama},
\textbf{Mistral~7B}~\citep{jiang2023mistral7b},
\textbf{OPT~6.7B}~\citep{zhang2022opt}, and
\textbf{Relu-LlaMA~7B}~\citep{sparsellm},
following GRIFFIN and TDA.
These models vary in architecture, training data, and optimization strategies, providing a representative set of contemporary foundation models.

For the LG task, noticing that most models above are not capable of generating cohesive long responses, we use the instruction-tuned pretrained models listed as follows: 
\textbf{Gemma~7B}~\citep{team2024gemma},
\textbf{Gemma~2~9B \& 27B}~\citep{gemma2},
\textbf{Mistral~7B}~\citep{jiang2023mistral7b},
\textbf{Qwen~2.5~7B \& 14B}~\citep{qwen2},
\textbf{Phi~3~14B~(medium)}~\citep{phi3}, 
\textbf{Llama~3~8B}~\citep{llama3}, and
\textbf{Gemma~3n~E2B \& E4B}. 
In particular, Gemma~3n models adopt MatFormer strategy~\citep{devvrit2024matformer} in training, making them inherently sparse and possibly more suitable for the application of inference time sparsification algorithms like GLASS. 

\paragraph{Evaluation Metrics.}
For evaluation of the C and SG tasks, following prior work, we use a set of evaluation metrics including classification accuracy, F1-Score, exact match, and ROUGE-1/2/L. 

For the LG benchmark, ROUGE or exact match can be misleading because small next-token distribution shifts compound over long generations. We therefore measure deviation at each step using Perplexity~(PPL) and Kullback-Leibler Divergence~(KLD)  (metrics details in App.~\ref{app:eval-metrics}) over next-token vocabulary distributions, conditioned on the dense model’s generation. 
\rC{Both are \emph{deviation-from-dense} metrics: the dense model has $\mathrm{KLD}\!=\!0$ by construction, while PPL measures how well the sparsified model preserves the dense model's generated trajectory.} 


\subsection{Performance Comparison on C and SG Tasks}

\rC{On Classification and Short-form Generation tasks, GLASS performs comparably to GRIFFIN, suggesting that global priors are most beneficial in the more challenging short-prompt/long-generation regime.} The largest improvement of GLASS over GRIFFIN was from \hbox{Mistral 7B} on the ARC-C benchmark, with a gain of 0.68\%. 
See Tab.~\ref{tab:sg} for details.
 
\subsection{Performance Comparison on Long Generation}
\label{sec:lg-results}
Tab.~\ref{tab:main} shows that both \hbox{A-GLASS} and \hbox{I-GLASS} improve over GRIFFIN across most models and metrics, with \hbox{I-GLASS} usually giving the strongest gains. For example, \hbox{I-GLASS} reduces PPL/KLD by 12.91\%/16.48\% on \hbox{Gemma 2 9B} and by 12.48\%/22.80\% on \hbox{Gemma 2 27B}. These results indicate that GLASS improves both next-token uncertainty and distributional fidelity across diverse model scales and architectures.

\textbf{MatFormer-style models.}
Largest gains appear on \hbox{Gemma 3n} family, which uses a MatFormer-style architecture designed for efficient on-device deployment~\citep{devvrit2024matformer}. Since MatFormer supports extracting smaller submodels from a larger trained model, these results suggest that GLASS is especially effective when structured sparsification is naturally supported. On \hbox{Gemma 3n E4B}, \hbox{A-GLASS}/\hbox{I-GLASS} improve PPL by 31.30\%/26.43\% and KLD by 25.26\%/22.25\%; on \hbox{Gemma 3n E2B}, the gains reach 45.10\%/37.72\% in PPL and 25.73\%/21.55\% in KLD.

\textbf{Qwen 2.5 7B exception.}
The main exception is \hbox{Qwen 2.5 7B}, where both variants slightly degrade PPL. We hypothesize this is due to a language mismatch: under NPS, the model tends to default to Chinese generation, while Alpaca is English, making the global prior partially misaligned with the evaluation distribution. Nevertheless, \hbox{I-GLASS} still improves KLD by 1.30\%, suggesting that the global prior retains some distributional value despite this mismatch.

\subsection{How Global Aggregation Enhances Local-Only Sparsification?}
\label{sec:how_ga}

To evaluate the effectiveness of aggregating global and local activation statistics, we run the following controlled experiment (details deferred to App.~\ref{app:how_ga}).


\textbf{Oracle-overlap analysis.}
To examine the mechanism behind GLASS, we perform a controlled oracle-overlap diagnostic on \hbox{Llama~3~8B}. We use two disjoint WikiText corpora to avoid evaluating on the same samples used to estimate \(A^g\): one corpus estimates global activation statistics, and the other provides a post hoc active-neuron oracle reference for overlap evaluation. At 50\% sparsity, \hbox{Global-Local} (\(A^l{+}A^g\)) produces masks that are substantially closer to this reference than either \hbox{Local-Only} (\(A^l\)) or \hbox{Global-Only} (\(A^g\)) alone. Specifically, as shown in Fig.~\ref{fig:jacc_radar} and Tab.~\ref{tab:jaccard-summary}, mean Jaccard increases from $0.527$~(\hbox{Local-Only}) and $0.510$~(\hbox{Global-Only}) to $0.604$, respectively, with lower layer-wise variance. These results provide direct evidence, within this diagnostic setting, that local and global signals are complementary: local activations capture input-specific salience, while global statistics stabilize the mask by preserving neurons that are consistently important across contexts.

\textbf{End-to-end PPL ablation.}
Tab.~\ref{tab:alpaca-ablation} in App.~\ref{app:how_ga} shows that better neuron selection translates to generation quality: \hbox{Global-Local} achieves the best PPL on all three models, improving over \hbox{Local-Only} by $8.19\%$, $23.28\%$, and $10.39\%$ on Gemma~7B, Llama~3~8B, and Mistral~7B, respectively. Together, these results show that GLASS benefits from synergistically fusing stable global importance with prompt-specific local evidence.

\subsection{Does NPS Produce More Reliable $I^{g}$ and $A^{g}$ for GLASS?}
\label{sec:nps-helps}
We evaluate whether NPS yields better global importance estimates by comparing it with WikiText-based global statistics across activation densities from 90\% to 10\% on \hbox{Gemma 7B}, \hbox{Llama 3 8B}, and \hbox{Mistral 7B} (Tab.~\ref{tab:kld_combined}). NPS-based GLASS consistently outperforms WikiText-based variants, indicating that NPS provides more reliable \(I^g\) and \(A^g\). The absolute KLD reductions over GRIFFIN are generally more pronounced in the low-density regime, suggesting that the global prior is especially useful under aggressive sparsification, where local-only masks are more likely to miss important neurons. \rA{For example, at 10\% density on \hbox{Llama 3 8B}, I-GLASS (NPS) reduces KLD from GRIFFIN's $5.1898$ to $4.4492$, a $14.27\%$ improvement.}

\begin{table}
\centering
\caption{KLD results across Gemma 7B, Llama 3 8B, and Mistral 7B.
Due to space limitation, GRIFFIN is denoted as GRFN and GLASS is denoted as GLS.
}
\scalebox{0.6125}{
\begin{tabular}{c|ccccc|ccccc|ccccc}
\toprule
& \multicolumn{5}{c|}{Gemma 7B} & \multicolumn{5}{c|}{Llama 3 8B} & \multicolumn{5}{c}{Mistral 7B} \\
\cmidrule(lr){2-6} \cmidrule(lr){7-11} \cmidrule(lr){12-16}
\shortstack{Density\\(\%)}
& \shortstack{GRFN\\\textcolor{white}{-}} & \shortstack{A-GLS\\(Wiki)} & \shortstack{A-GLS\\(NPS)} & \shortstack{I-GLS\\(Wiki)} & \shortstack{I-GLS\\(NPS)}
& \shortstack{GRFN\\\textcolor{white}{-}} & \shortstack{A-GLS\\(Wiki)} & \shortstack{A-GLS\\(NPS)} & \shortstack{I-GLS\\(Wiki)} & \shortstack{I-GLS\\(NPS)}
& \shortstack{GRFN\\\textcolor{white}{-}} & \shortstack{A-GLS\\(Wiki)} & \shortstack{A-GLS\\(NPS)} & \shortstack{I-GLS\\(Wiki)} & \shortstack{I-GLS\\(NPS)} \\
\midrule
90 & 0.0578 & 0.0529 & 0.0392 & 0.0467 & 0.0347 & 0.0760 & 0.0690 & 0.0534 & 0.0627 & 0.0505 & 0.0965 & 0.0876 & 0.0694 & 0.0807 & 0.0648 \\
80 & 0.1453 & 0.1458 & 0.1127 & 0.1324 & 0.1025 & 0.1904 & 0.1795 & 0.1438 & 0.1699 & 0.1369 & 0.2170 & 0.2098 & 0.1755 & 0.1985 & 0.1669 \\
70 & 0.2659 & 0.2782 & 0.2243 & 0.2593 & 0.2076 & 0.3298 & 0.3237 & 0.2739 & 0.3168 & 0.2644 & 0.3638 & 0.3593 & 0.3124 & 0.3460 & 0.3031 \\
60 & 0.4266 & 0.4563 & 0.3799 & 0.4359 & 0.3552 & 0.5207 & 0.5099 & 0.4492 & 0.5022 & 0.4434 & 0.5650 & 0.5532 & 0.4971 & 0.5419 & 0.4906 \\
50 & 0.6453 & 0.7022 & 0.5998 & 0.6825 & 0.5661 & 0.9426 & 0.7748 & 0.7407 & 0.7646 & 0.7354 & 0.8774 & 0.8316 & 0.7615 & 0.8300 & 0.7593 \\
40 & 0.9735 & 1.1184 & 0.9610 & 1.1091 & 0.9107 & 1.3079 & 1.1942 & 1.0902 & 1.1801 & 1.0702 & 1.3697 & 1.2614 & 1.1570 & 1.2795 & 1.1575 \\
30 & 1.6049 & 1.8326 & 1.6001 & 1.8205 & 1.5179 & 2.0843 & 1.7930 & 1.6937 & 1.7797 & 1.6746 & 2.1444 & 1.8603 & 1.7323 & 1.9065 & 1.7429 \\
20 & 3.2728 & 3.5411 & 3.1478 & 3.5111 & 2.9906 & 3.3302 & 2.8017 & 2.7080 & 2.7759 & 2.6770 & 3.3806 & 2.8508 & 2.6695 & 2.9881 & 2.7417 \\
10 & 8.3718 & 8.5014 & 7.7850 & 8.1909 & 7.4167 & 5.1898 & 5.1234 & 4.3153 & 5.0352 & 4.4492 & 5.2786 & 5.2224 & 4.3848 & 5.1669 & 4.5143 \\
\bottomrule
\end{tabular}
}
\label{tab:kld_combined}
\end{table}

\subsection{On-Device Runtime Performance}
\label{sec:on-device}

We evaluate GLASS with 50\% FFN static masking applied during the decoding phase on a Samsung \hbox{Galaxy~S25~Ultra}, using \hbox{Qwen3 4B}, \hbox{Llama3 8B}, and \hbox{Gemma 7B}. 
We run all on-device experiments using the LiteRT-LM library~\citep{litertlm} and convert models using the LiteRT-Torch~\citep{literttorch}.
All experiments are averaged over 30 runs with a fixed prefill length of 1024 tokens for \hbox{Qwen3 4B} and \hbox{Llama3 8B}, and 64 for \hbox{Gemma 7B}. 
GLASS achieves an average decoding speedup of 
20\% for \hbox{Qwen3 4B} when generating 
256 tokens.
For \hbox{Llama3 8B}, the corresponding improvement is 42\% relative to the dense baseline. 
In the \hbox{Gemma 7B} case, the 50\% FFN reduction enables full RAM residency on device, yielding $\sim\!11\times$ decoding speedup.
App.~\ref{additional_edge} presents the details and runtime comparisons.

\section{Related Works}
\label{sec:prev_work}

\paragraph{Inference-time sparsification of LLMs.}
Inference-time sparsification accelerates LLM decoding by exploiting the inherent sparsity of FFN activations, complementing traditional static pruning. Existing methods can be broadly grouped into training-based and training-free approaches. Training-based methods such as DejaVu~\citep{liu2023deja}, ShadowLLM~\citep{akhauri2024shadowllm}, and PowerInfer~\citep{song2024powerinfer,xue2024powerinfer} rely on learned predictors or auxiliary models to decide which components to execute at inference time. In contrast, training-free methods avoid additional model training. GRIFFIN~\citep{dong2024prompt} summarizes FFN activation magnitudes during prefill and reuses the resulting mask during decoding. TDA~\citep{ma2024first} uses prefill activations to set layer-wise activation thresholds, while TEAL~\citep{liu2024teal} greedily optimizes layer-wise sparsity levels. CATS~\citep{lee2404cats} instead estimates thresholds from offline activation statistics. GLASS is closest to GRIFFIN in that it is training-free and uses a fixed prefill-time mask, but differs by augmenting prompt-local activation evidence with a global model-intrinsic prior.

\paragraph{Static versus dynamic masks for edge deployment.}
Beyond the training-free/training-based distinction, another key deployment axis is whether masks are \emph{static} or \emph{dynamic}. GLASS and GRIFFIN compute a fixed prefill-time mask and reuse it during decoding, avoiding per-token mask recomputation and enabling a compact subset of FFN weights to remain resident in fast memory. In contrast, DejaVu and TEAL use dynamic masks, which introduce per-token overhead and frequent weight movement, making edge deployment more challenging.


DejaVu trains per-layer predictors and recomputes active neurons at every decoding step, adding predictor overhead and preventing fixed weight residency. TEAL addresses an orthogonal problem: it determines \emph{how much} to prune per layer, whereas GLASS determines \emph{which} neurons to keep under a given sparsity budget. Thus, TEAL's layer-wise allocation could in principle be combined with GLASS's neuron selection. GRIFFIN is the most direct baseline because it is both training-free and static, but it relies only on prompt-derived local evidence; GLASS improves robustness by fusing this signal with a global prior.

\textbf{Neuron Importance Estimation.} The pursuit of identifying and ranking neuron importance has been a core focus of neural network research since the field's inception.
{Optimal Brain Damage~(OBD)}~\citep{lecun1989optimal} and {Optimal Brain Surgeon~(OBS)}~\citep{hassibi1993optimal}
are among the earliest approaches to network pruning by estimating the impact of individual parameters on training loss. 
They use a second-order Taylor expansion of the loss with respect to \emph{model~weights}, 
and compute saliency scores using the diagonal~(OBD) or full~(OBS) Hessian of the loss. 
While theoretically grounded, these methods are computationally expensive and scale poorly with model size, limiting their applicability to modern large-scale networks.
{PerforatedCNNs}~\citep{figurnov2016perforatedcnns} apply first-order Taylor expansion for structured pruning, focusing on reducing spatial redundancy in convolutional neural networks. 
Similarly, \citet{molchanov2016pruning} apply the first-order Taylor criterion to prune convolutional channels. \citet{molchanov2019importance} generalizes this to second-order expansions, and \citet{kwon2022fast} introduce a Fisher information-based, training-free pruning framework for Transformers, enhancing inference efficiency.


\paragraph{Null Prompt Stimulation.}
Null or empty prompts have been used to probe LLM behavior~\citep{jeong2025scope,leidinger2023language}. GLASS uses them for a different purpose: estimating model-intrinsic neuron importance. By letting the model generate its own stimulation data, NPS avoids reliance on a task dataset or external corpus, reducing potential corpus bias and preserving the plug-and-play nature of GLASS. To our knowledge, GLASS is the first method we are aware of that uses null-prompt stimulation for FFN neuron-importance estimation.

\paragraph{Conditional computation, sparse attention, and test-time adaptation.}
Conditional computation adaptively allocates computation based on the input. MoE models~\citep{shazeer2017outrageously} route tokens to selected experts, SkipGPT~\citep{zhaoskipgpt} performs token-aware layer skipping, and MatFormer~\citep{devvrit2024matformer} trains nested submodels that can be extracted at inference time. These methods require architectural or training-time design, whereas GLASS is applied post hoc to pretrained LLMs. Sparse attention~\citep{child2019generating,yuan2025native} is orthogonal because it accelerates attention blocks rather than FFN layers, and can in principle be combined with GLASS. Test-time training adapts model weights at inference time; GLASS instead keeps weights fixed and adapts only the FFN sparsity mask using local and global neuron-importance estimates.

\section{Conclusion} \label{sec:conclusion}
In this paper, we introduced two variants of a training-free method for inference-time sparsification of FFNs in LLMs. 
By fusing local prompt activations with global statistics, either activation magnitudes~(\hbox{A-GLASS}) or impact scores~(\hbox{I-GLASS}), 
our methods strike a balance between context sensitivity and model-intrinsic diversity without requiring offline predictor training or incurring runtime overhead. \rAS{We also showed that this rank-aggregation rule is not ad hoc, but admits a probabilistic interpretation: under a Mallows-type model with squared Spearman distance, it corresponds to the MAP consensus ranking.}


Extensive experiments across multiple tasks and models demonstrate that our methods 
outperform prior training-free approaches in the vast majority of cases, especially under short prompts and long generations. 
These results demonstrate that model-intrinsic global knowledge can significantly improve prompt-driven sparsification, unlocking reliable speedups even for the short-prompt long-generation workloads typical of edge devices. Our work opens the door for efficient, adaptive inference in LLMs, which is particularly important for resource-constrained edge devices.

\paragraph{Limitations and Future Work.}
Building on the strong performance of GLASS, we identify a few opportunities to further improve its flexibility and effectiveness:
(i)~currently we apply a fixed sparsity level uniformly and share a single mixing coefficient \(\lambda\) across all layers; jointly optimizing the sparsity pattern and learning layer-specific \(\lambda\) values could lead to more efficient capacity allocation; (ii)~each layer is pruned independently for simplicity and compatibility with standard sparsification frameworks, however, this may neglect potential cross-layer interactions that coordinated pruning strategies could leverage.  

\bibliographystyle{abbrvnat}
\bibliography{references}


\appendix
\newpage
\appendix
\onecolumn

\section{Permutation-Based Consensus View of GLASS}
\label{app:glass-consensus-proof}

For completeness, we provide the full derivation of the MAP consensus rule used in Section~\ref{sec:glass:formulation}. Let $\pi$ denote the latent consensus permutation over $m$ neurons. Let $\pi^{(l)}$ and $\pi^{(g)}$ denote the observed local and global permutations induced by sorting the local and global importance scores, respectively. The permutation $\pi=(\pi_1,\dots,\pi_m)$ lists neurons in ranked order from least important to most important, while the rank vector $r(\pi)$ records, for each neuron index $j$, the rank position assigned to $j$. Under this convention, rank $1$ corresponds to the least important neuron and rank $m$ corresponds to the most important neuron. Equivalently, $r(\pi)$ is the inverse permutation written in vector form. \rAS{If exact ties occur in the local or global importance scores before rank conversion, we use stable deterministic tie-breaking by neuron index to obtain a valid permutation. In practice, exact ties are rare because $A^l$, $A^g$, and $I^g$ are floating-point averages over many tokens. If a tie occurs at the top-$k$ boundary, the same deterministic rule is used to ensure reproducible neuron selection.}

We assume the following conditional ranking models:
\begin{equation}
P(\pi^{(l)} \mid \pi)
=
\frac{1}{Z_l}
\exp\!\left(
-\beta_l \, \|r(\pi^{(l)}) - r(\pi)\|_2^2
\right),
\end{equation}
\begin{equation}
P(\pi^{(g)} \mid \pi)
=
\frac{1}{Z_g}
\exp\!\left(
-\beta_g \, \|r(\pi^{(g)}) - r(\pi)\|_2^2
\right),
\end{equation}
where $\beta_l,\beta_g>0$ are concentration parameters controlling the reliability of the local and global rankings. \rT{The squared rank distance $\|r(\sigma_1)-r(\sigma_2)\|_2^2 = \sum_j (R_j^{(\sigma_1)} - R_j^{(\sigma_2)})^2$ is the standard \emph{squared Spearman rank distance} (Spearman's $\rho$-distance), and the resulting Mallows-type model is well-defined on the space of permutations~\citep{mallows1957non,diaconis1977spearman,fligner1986distance}.} \rAS{We use squared Spearman distance because GLASS operates on rank-position vectors and because this distance penalizes the magnitude of rank displacement, yielding a closed-form MAP consensus ranking via weighted rank sums.} For MAP estimation, the normalization constants do not affect the optimizer and can be dropped. 

\rAS{We also assume a uniform prior over the latent consensus permutation, $P(\pi)=1/m!$. Therefore,
\[
P(\pi \mid \pi^{(l)},\pi^{(g)})
\propto
P(\pi^{(l)},\pi^{(g)} \mid \pi),
\]
so MAP estimation is equivalent to maximizing the likelihood.}

\rAS{As a simplifying modeling assumption, we assume conditional independence of the two observed rankings given the latent consensus permutation:}
\begin{equation}
P(\pi^{(l)},\pi^{(g)} \mid \pi)
=
P(\pi^{(l)} \mid \pi)\,P(\pi^{(g)} \mid \pi).
\end{equation}
Substituting the two likelihoods,
\begin{equation}
P(\pi^{(l)},\pi^{(g)} \mid \pi)
\propto
\exp\!\left(
-\beta_l \|r(\pi^{(l)}) - r(\pi)\|_2^2
-\beta_g \|r(\pi^{(g)}) - r(\pi)\|_2^2
\right).
\end{equation}

The MAP estimator is
\begin{equation}
\pi^\star
=
\arg\max_{\pi}
P(\pi^{(l)},\pi^{(g)} \mid \pi).
\end{equation}
Taking logarithms and dropping additive constants independent of $\pi$ gives
\begin{equation}
\pi^\star
=
\arg\min_{\pi}
\beta_l \|r(\pi^{(l)}) - r(\pi)\|_2^2
+
\beta_g \|r(\pi^{(g)}) - r(\pi)\|_2^2.
\label{eq:app-map-objective}
\end{equation}

For shorthand, define
\begin{equation}
r := r(\pi), \qquad r_l := r(\pi^{(l)}), \qquad r_g := r(\pi^{(g)}).
\end{equation}
Then Eq.~\eqref{eq:app-map-objective} becomes
\begin{equation}
\pi^\star
=
\arg\min_{r \in \mathcal{P}}
\beta_l \|r-r_l\|_2^2 + \beta_g \|r-r_g\|_2^2,
\end{equation}
where $\mathcal{P}$ denotes the set of all rank vectors corresponding to permutations of $(1,\dots,m)$.

Expanding the squared norms,
\begin{equation}
\|r-r_l\|_2^2
=
\|r\|_2^2 - 2r^\top r_l + \|r_l\|_2^2,
\end{equation}
\begin{equation}
\|r-r_g\|_2^2
=
\|r\|_2^2 - 2r^\top r_g + \|r_g\|_2^2.
\end{equation}
Therefore,
\begin{align}
\beta_l \|r-r_l\|_2^2 + \beta_g \|r-r_g\|_2^2
&=
\beta_l \bigl(\|r\|_2^2 - 2r^\top r_l + \|r_l\|_2^2\bigr)
+
\beta_g \bigl(\|r\|_2^2 - 2r^\top r_g + \|r_g\|_2^2\bigr) \\
&=
(\beta_l+\beta_g)\|r\|_2^2
-2r^\top(\beta_l r_l+\beta_g r_g)
+
\beta_l\|r_l\|_2^2+
\beta_g\|r_g\|_2^2.
\label{eq:app-expanded-map}
\end{align}
\rT{The crucial observation is now: if $r$ is a rank vector corresponding to some permutation, then it is always a permutation of $(1,2,\dots,m)$. Hence}
\begin{equation}
\|r\|_2^2 = \sum_{k=1}^m k^2 = \frac{m(m+1)(2m+1)}{6},
\end{equation}
\rAS{which is constant over all feasible $r\in\mathcal{P}$. Moreover, since $r_l$ and $r_g$ are also rank vectors corresponding to permutations of $(1,\dots,m)$, they have the same squared norm:
\[
\|r_l\|_2^2=\|r_g\|_2^2=\sum_{k=1}^m k^2.
\]
Since $r_l$ and $r_g$ are fixed observations, all norm terms in Eq.~\eqref{eq:app-expanded-map} are constant with respect to the candidate consensus permutation. Therefore, the MAP problem is equivalent to}
\begin{equation}
\pi^\star
=
\arg\max_{r \in \mathcal{P}}
r^\top(\beta_l r_l+\beta_g r_g).
\label{eq:app-linear-map}
\end{equation}

Define the score vector
\begin{equation}
s := \beta_l r_l + \beta_g r_g.
\end{equation}
Then Eq.~\eqref{eq:app-linear-map} becomes
\begin{equation}
\pi^\star
=
\arg\max_{r \in \mathcal{P}} r^\top s.
\end{equation}
\rT{Since $r$ must be a permutation of $(1,\dots,m)$, $r^\top s = \sum_{j=1}^m r_j s_j$ is a sum where the values $\{1,2,\dots,m\}$ are distributed among the indices $j$, weighted by the fixed entries $s_j$. By the rearrangement inequality~\citep{hardy1952inequalities}, this sum is maximized when the largest value $m$ is paired with the index $j^\star$ such that $s_{j^\star}$ is maximal, the second-largest value $m-1$ is paired with the index achieving the second-largest $s$, and so on. Equivalently, the optimal permutation is obtained by sorting indices by $s_j$ in descending order and assigning rank $m$ to the index with the largest score, $m-1$ to the next, etc.} Under our convention, larger rank values indicate higher importance. Therefore, the consensus ordering from most important to least important is obtained by sorting neurons in descending order of $s_j$. Writing
\begin{equation}
R_j^{(l)} := [r_l]_j, \qquad R_j^{(g)} := [r_g]_j,
\end{equation}
we have
\begin{equation}
s_j = \beta_l R_j^{(l)} + \beta_g R_j^{(g)}.
\end{equation}
Finally, normalizing by $\beta_l+\beta_g$ does not change the induced ordering. Let
\begin{equation}
\lambda = \frac{\beta_g}{\beta_l+\beta_g}.
\end{equation}
Then
\begin{equation}
\frac{s_j}{\beta_l+\beta_g}
=
(1-\lambda) R_j^{(l)} + \lambda R_j^{(g)}.
\end{equation}
Since multiplication by a positive constant preserves ordering, the same MAP permutation is obtained by sorting in descending order according to
\begin{equation}
(1-\lambda) R_j^{(l)} + \lambda R_j^{(g)}.
\end{equation}
Thus, under the convention where larger rank values indicate higher importance, the GLASS score arises as the MAP consensus score under this Mallows-type ranking model with squared Spearman rank distance. In practice, GLASS only needs the $k$ neurons with the largest GLASS scores under this induced ranking, not the entire permutation.

\rAS{\paragraph{Remark on uniqueness and ties.}
The argmax is unique whenever all $s_j$ are distinct. If ties occur in the combined score $s_j$, multiple consensus permutations achieve the same MAP objective value, and any consistent tie-breaking rule recovers a valid optimum. This is separate from ties in the original local or global importance scores before rank conversion, which we handle by stable deterministic tie-breaking as described above.}

\section{Additional Experiment Details}

\subsection{Benchmark Details}
\label{app:benchmark-details}
\subsubsection{Classification Benchmark}
\textbf{HellaSwag}~\citep{zellers2019hellaswag} is a commonsense inference dataset where the task is to choose the most plausible continuation of a given context from four choices.
\textbf{PIQA}~\citep{Bisk2020} is a physical commonsense reasoning dataset focused on selecting the most feasible solution to everyday tasks involving physical interactions.
\textbf{COPA}~\citep{roemmele2011choice} is a causal reasoning dataset where models must choose the more plausible cause or effect of a given premise.
\textbf{ARC-Easy/Challenge}~\citep{allenai:arc} is a benchmark for grade-school science question answering, with Easy posing straightforward factual questions and Challenge requiring complex reasoning.
\textbf{BoolQ}~\citep{clark2019boolq} is a question-answering dataset consisting of yes/no questions paired with paragraphs from Wikipedia, requiring models to determine the correct answer based on the context.

\subsubsection{Short-form Generation Benchmark}
\textbf{XSum}~\citep{Narayan2018DontGM} is a dataset for extreme summarization, where the goal is to generate a single-sentence summary capturing the key information from a BBC news article.
\textbf{CNN/DailyMail}~\citep{see-etal-2017-get,DBLP:conf/nips/HermannKGEKSB15} is a large-scale dataset for abstractive summarization, consisting of news articles paired with multi-sentence highlights as summaries.
\textbf{CoQA}~\citep{reddy-etal-2019-coqa} is a conversational question answering dataset where models answer a sequence of interconnected questions grounded in a passage.
\textbf{QASPER}~\citep{dasigi2021dataset} is a dataset of question-answer pairs over scientific papers, designed for training and evaluating systems on long-context, evidence-based QA.

\subsection{Details of Evaluation Metrics}
\label{app:eval-metrics}
\subsubsection{Perplexity (PPL)}
Perplexity (PPL) gauges how surprised a language model is when predicting the next token in a sequence.
It is the exponentiated average negative log-likelihood per token:
\begin{equation}
    \text{PPL} = \exp\Bigl(-\frac{1}{N}\sum_{i=1}^{N}\log p_\theta(x_i)\Bigr),
\end{equation}
where \(x_i\) is the \(i\)-th token and \(p_\theta(x_i)\) is the model's output probability density under parameters \(\theta\). 
Lower values indicate better predictive performance. 

We use the unpruned model's generated response as the reference for computing PPL, which measures how unlikely the dense model's generated trajectory is under the sparsified model. This protocol evaluates how well the sparsified model preserves the dense model's chosen continuation. Values reported in the main tables therefore quantify how much harder the sparsified model finds the dense model's tokens compared with the dense reference trajectory.


\subsubsection{Kullback-Leibler Divergence (KLD)}
The average token-level Kullback-Leibler Divergence (KLD) between the reference distribution \(P_i\) and the model distribution \(Q_i\) is
\begin{equation}    
    \text{KLD} \;=\; \frac{1}{N}\sum_{i=1}^{N}
          \sum_{v\in\mathcal V} P_i(v)\,\log \frac{P_i(v)}{Q_i(v)},
\end{equation}
where \(N\) is the sequence length, \(v \in \mathcal{V}\) the vocabulary,
\(P_i\) the reference token distribution, and
\(Q_i = \mathrm{softmax}(\text{logits}_i)\) is the model-predicted distribution
for the \(i\)-th position. \rC{When the sparsified model coincides with the dense reference, $P_i = Q_i$ and $\mathrm{KLD}=0$ exactly, so the values reported in Tab.~\ref{tab:main} directly quantify deviation from the dense baseline.}

\rAS{Since the vocabulary is very large, we report an approximate top-100 KLD computed over the 100 tokens with highest probability under the unpruned model. We restrict both distributions to this token set and renormalize them before computing KL. Empirically, this subset captures almost all the probability mass, with their summed probabilities effectively equal to $1$ in our evaluated settings.}

\subsubsection{Classification Accuracy}
Classification accuracy is the share of predictions that exactly match the ground-truth label:
\[
\text{Accuracy} \;=\; \frac{1}{N}\,\sum_{i=1}^{N}\mathbf 1\bigl\{\,\hat y_i = y_i\,\bigr\},
\]
where \(N\) is the number of examples, \(y_i\) is the reference label, \(\hat y_i\) the model's label, and \(\mathbf 1\{\cdot\}\) the indicator function.

\subsubsection{ROUGE-1 / 2 / L}
ROUGE (Recall-Oriented Understudy for Gisting Evaluation) compares a system output to one or more reference texts.
\begin{itemize}
  \item \textbf{ROUGE-\(n\)} (\(n=1,2\)) measures \(n\)-gram recall:
    \[
      \text{ROUGE-}n = 
      \frac{\sum_{g\in\mathrm{Ref}_n}\min\bigl(\mathrm{Count}_{\text{Hyp}}(g),\,\mathrm{Count}_{\text{Ref}}(g)\bigr)}
           {\sum_{g\in\mathrm{Ref}_n}\mathrm{Count}_{\text{Ref}}(g)}.
    \]
  \item \textbf{ROUGE-L} uses the longest common subsequence (LCS) of length \(L_{\text{LCS}}\):
    \[
      \text{ROUGE-L} = \frac{(1+\beta^{2})\,R_{\text{LCS}}\,P_{\text{LCS}}}{\beta^{2}\,P_{\text{LCS}} + R_{\text{LCS}}}
    \]
    \[
      R_{\text{LCS}} = \frac{L_{\text{LCS}}}{|\text{Ref}|}, 
      \qquad
      P_{\text{LCS}} = \frac{L_{\text{LCS}}}{|\text{Hyp}|}.
    \]
\end{itemize}

\subsubsection{Exact Match}
Exact Match (EM) is the proportion of predictions whose normalized string (e.g., lower-cased, stripped of punctuation and articles) matches the reference answer exactly:
\[
\text{EM} = \frac{1}{N}\sum_{i=1}^{N}\mathbf 1\bigl\{\mathrm{norm}(\hat a_i)=\mathrm{norm}(a_i)\bigr\}.
\]

\subsubsection{F1-Score}
In LLM evaluation (e.g., extractive QA or short-form generation), the F1-score is computed at the token level between the model's answer \(\hat a\) and the reference answer \(a\).

\medskip
After normalization (lower-casing, removing punctuation and articles) both answers are tokenized into multisets \(T_{\text{pred}}\) and \(T_{\text{ref}}\).  
Let \(C = \lvert\,T_{\text{pred}} \cap T_{\text{ref}}\,\rvert\) be the size of their multiset intersection. Then
\[
P = \frac{C}{\lvert T_{\text{pred}}\rvert},\qquad
R = \frac{C}{\lvert T_{\text{ref}}\rvert},\qquad
F_1 = 2\,\frac{P\,R}{P + R}.
\]
For a dataset with \(N\) examples, the overall score is the mean of the per-example values:
\[
\text{F1}_{\text{avg}} = \frac{1}{N}\sum_{i=1}^{N} F_{1}^{(i)}.
\]

\subsection{Implementation Details}
\label{app:imp_detail}
\rAS{Tab.~\ref{tab:nps-implementation-details} summarizes the implementation details for the global-prior computation. For A-GLASS, we only collect normalized activation magnitudes during forward passes. For I-GLASS, we first generate NPS sequences from the dense model, then replay each generated sequence with teacher forcing and use the self-generated next token as the pseudo-label for the cross-entropy loss. We accumulate $|h_j(x)\delta_j(x)|$ over tokens and sequences to obtain one global impact vector per layer. These global vectors are stored once per model and reused for all downstream prompts.}
\label{app:imp-det}


\paragraph{NPS.}
Since some pretrained models tend to repeat the same token or sequence of tokens when given minimal conditioning, we applied NPS as follows to encourage diversity in generation: For the first 10 tokens, we set the temperature to 1.5 and enabled a bigram repetition penalty to maximize initial diversity. After that, we reduced the temperature to 1 and disabled the penalty to avoid generating out-of-distribution text. Here are the rest of the generation parameters: number of samples = 1000, sequence length = 1024, top-k = 20.

\paragraph{Compute.}
All experiments were performed on a single NVIDIA GPU with 80GB memory. \rAS{\hbox{I-GLASS} is more expensive to compute offline than A-GLASS because it requires backward passes for gradient collection. However, this cost is incurred only once per model and is not paid during inference: the resulting global importance vectors are stored and reused for all prompts and deployments. At deployment time, both A-GLASS and I-GLASS only require rank fusion with local prefill statistics and application of the resulting FFN mask. We report the required passes and hardware rather than a single wall-clock precomputation time because offline runtime depends strongly on model size and implementation details.}

\begin{table}[t]
\centering
\caption{Implementation details for A-GLASS and I-GLASS under NPS.}
\label{tab:nps-implementation-details}
\begin{tabular}{lcc}
\toprule
\textbf{Item} & \textbf{A-GLASS} & \textbf{I-GLASS} \\
\midrule
NPS sequences & 1000 & 1000 \\
Max sequence length & 1024 & 1024 \\
Target for gradients & N/A & Self-generated next token \\
Passes & Forward only & Forward + backward \\
GPU & 1$\times$80GB & 1$\times$80GB \\
Precision & bf16 & bf16 \\
Stored statistic & Per-layer vector $A^g$ & Per-layer vector $I^g$ \\
\bottomrule
\end{tabular}
\end{table}

\section{Additional Results and Discussions}
\subsection{How Global Aggregation Enhances Local-Only Sparsification?}
\label{app:how_ga}

We provide two controlled analyses to isolate the contribution of global-local aggregation: an oracle-overlap analysis, which tests \emph{which} neurons are selected, and an end-to-end PPL ablation, which tests \emph{how} the selected masks affect generation quality.

\paragraph{Oracle-overlap analysis.}
Using WikiText~\citep{merity2016pointer}, we construct two disjoint corpora, each with 100 sequences of exactly 1024 tokens, formed by concatenating shorter documents. The first corpus is used to compute global activation statistics \(A^{g}\). On the second corpus, we compare three 50\%-sparsity variants:
(i)~\textbf{Local-Only}, which ranks neurons using only per-input activations \(A^{l}\);
(ii)~\textbf{Global-Only}, which uses only \(A^{g}\); and
(iii)~\textbf{Global-Local}, which aggregates \(A^{g}\) and \(A^{l}\).

We evaluate each variant by its Jaccard similarity to an oracle critical-neuron set. Oracle critical neurons are the top-50\% neurons by post-hoc decoding-time activation magnitude for each input. Although unavailable to practical dynamic pruning methods, this oracle provides a useful controlled reference. For sets \(A\) and \(B\), the Jaccard similarity is
\[
J(A,B)=\frac{|A\cap B|}{|A\cup B|}.
\]

On Llama~3~8B, Global-Local consistently better matches the oracle than either Local-Only or Global-Only. Aggregated across all 32 layers, Global-Local reaches a mean Jaccard of $0.604 \pm 0.048$, compared to $0.527 \pm 0.066$ for Local-Only and $0.510 \pm 0.082$ for Global-Only. Thus, fusion improves both accuracy and consistency: it lifts the two individual signals by $14$--$18\%$ and reduces layer-wise variance, indicating synergy rather than dominance by either signal.

\begin{table}[t]
\centering
\caption{Layer-aggregated summary of the oracle-overlap analysis in Fig.~\ref{fig:jacc_radar}. We report Jaccard similarity to the oracle on Llama 3 8B, mean and std across all 32 layers, at 50\% sparsity.}
\label{tab:jaccard-summary}
\small
\begin{tabular}{lcc}
\toprule
\textbf{Variant} & \textbf{Mean Jaccard} & \textbf{Std} \\
\midrule
Local-Only      & 0.527 & 0.066 \\
Global-Only     & 0.510 & 0.082 \\
\textbf{Global-Local (Ours)} & \textbf{0.604} & \textbf{0.048} \\
\bottomrule
\end{tabular}
\end{table}
\paragraph{End-to-end PPL ablation.}
We next test whether improved oracle overlap translates into generation quality on the Alpaca short-prompt/long-generation benchmark. Tab.~\ref{tab:alpaca-ablation} reports PPL at 50\% sparsity for Gemma~7B, Llama~3~8B, and Mistral~7B.

\begin{table}[t]
\centering
\caption{PPL on the Alpaca short-prompt/long-generation benchmark at 50\% FFN sparsity. Numbers in parentheses are standard deviations across samples.}
\label{tab:alpaca-ablation}
\small
\begin{tabular}{lccc}
\toprule
\textbf{Variant} & \textbf{Gemma 7B} & \textbf{Llama 3 8B} & \textbf{Mistral 7B} \\
\midrule
Local-Only \scriptsize{($\lambda{=}0$; GRIFFIN)}
& 3.7014 \scriptsize{(0.0316)}
& 5.2523 \scriptsize{(0.0434)}
& 5.0059 \scriptsize{(0.0407)} \\
Global-Only \scriptsize{($\lambda{=}1$; static global mask)}
& 5.6611 \scriptsize{(6.5708)}
& 4.6872 \scriptsize{(2.5860)}
& 6.3699 \scriptsize{(3.6886)} \\
\textbf{Global+Local} \scriptsize{($\lambda{=}0.5$; I-GLASS)}
& \textbf{3.3982} \scriptsize{(0.0310)}
& \textbf{4.0295} \scriptsize{(0.0331)}
& \textbf{4.4860} \scriptsize{(0.0387)} \\
\bottomrule
\end{tabular}
\end{table}

Two conclusions follow. First, Global-Only is not a reliable standalone strategy: it has substantially higher variance than the prompt-conditioned methods, indicating that a static input-agnostic mask is unstable across diverse prompts. Second, Global+Local achieves the best mean PPL on all three models, improving over Local-Only by $8.19\%$, $23.28\%$, and $10.39\%$, respectively. Since Global-Only alone is often worse than Local-Only, the gains cannot be explained by the global prior alone; they arise from fusing stable model-intrinsic importance with prompt-specific evidence.

\rA{\subsection{$\lambda$ Sensitivity Sweep}
\label{app:lambda-sweep}

\begin{figure}[b]
  \centering
  \includegraphics[width=0.6\columnwidth]{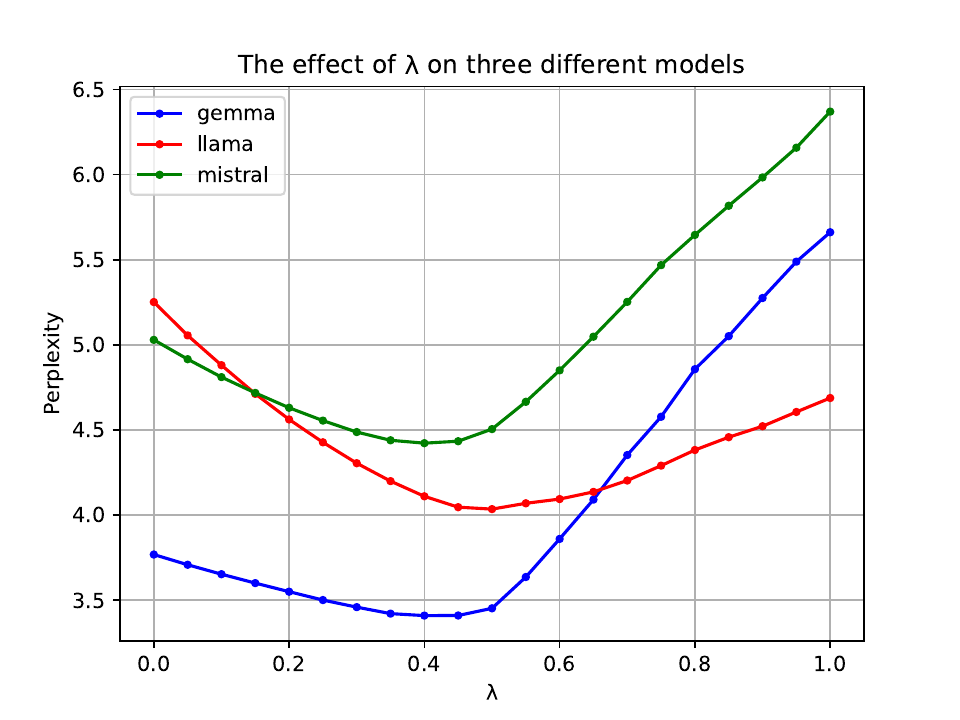}
  \caption{\rA{Sensitivity of PPL to the mixing weight $\lambda$ at 50\% sparsity. The minima cluster tightly around $\lambda=0.5$ across architectures, supporting the use of an untuned default.}}
  \label{fig:lambda-sweep}
\end{figure}}

We performed a fine-grained sweep over $\lambda\in[0,1]$ with step size $0.05$ on three representative instruction-tuned models (Gemma 7B, Llama 3 8B, Mistral 7B) at 50\% sparsity on the Alpaca long-generation benchmark using I-GLASS with NPS. For each $\lambda$, we report mean PPL across the same 3{,}602 samples used in Tab.~\ref{tab:main}. Two findings stand out:
\begin{itemize}
  \item The PPL landscape is \emph{smooth and unimodal} for all three models, with no abrupt transitions. This indicates that the rank-aggregation rule degrades gracefully as $\lambda$ moves away from its optimum.
  \item All three models attain their best PPL near $\lambda=0.5$. The optima do not require model-specific tuning, supporting the principled choice of $\lambda=0.5$ as the equal-reliability default ($\beta_l=\beta_g$ in the Mallows model of Sec.~\ref{sec:glass:formulation}).
\end{itemize}
The two endpoints recover the natural baselines: $\lambda=0$ corresponds to GRIFFIN (Local-Only) and $\lambda=1$ corresponds to a static global mask (Global-Only). Both are \emph{worse} than the fused $\lambda\approx 0.5$ on every model evaluated, in agreement with the Tab.~\ref{tab:alpaca-ablation} ablation in the main text. We provide the full sweep in Fig.~\ref{fig:lambda-sweep}.

\subsection{Edge Device Experiments} \label{additional_edge}
\begin{figure}[htbp]
\centering

\begin{subfigure}{0.45\textwidth}
    \centering
    \includegraphics[width=\linewidth]{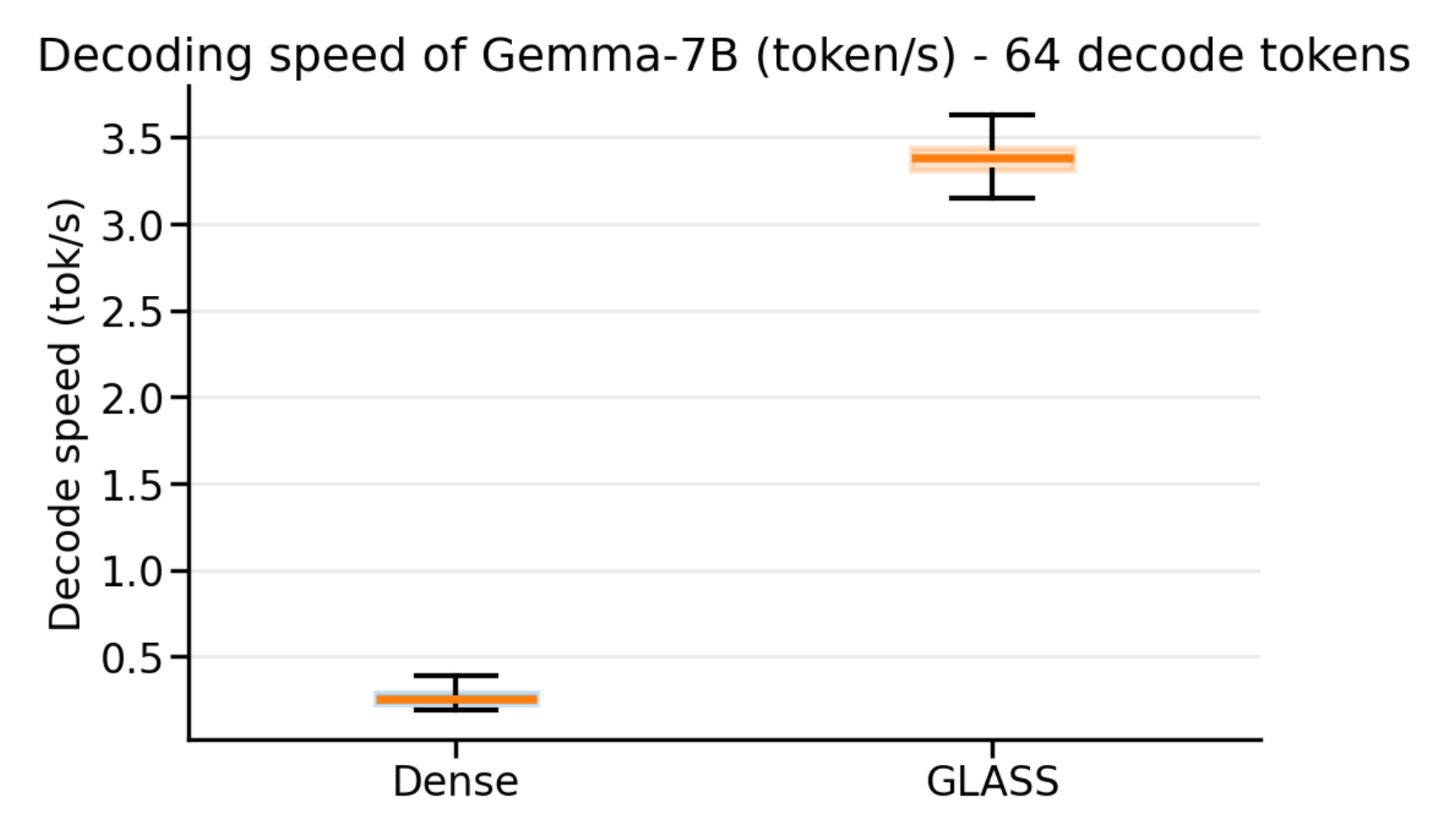}
    \caption{ }
    \label{fig:speedup_64}
\end{subfigure}
\hfill
\begin{subfigure}{0.45\textwidth}
    \centering
    \includegraphics[width=\linewidth]{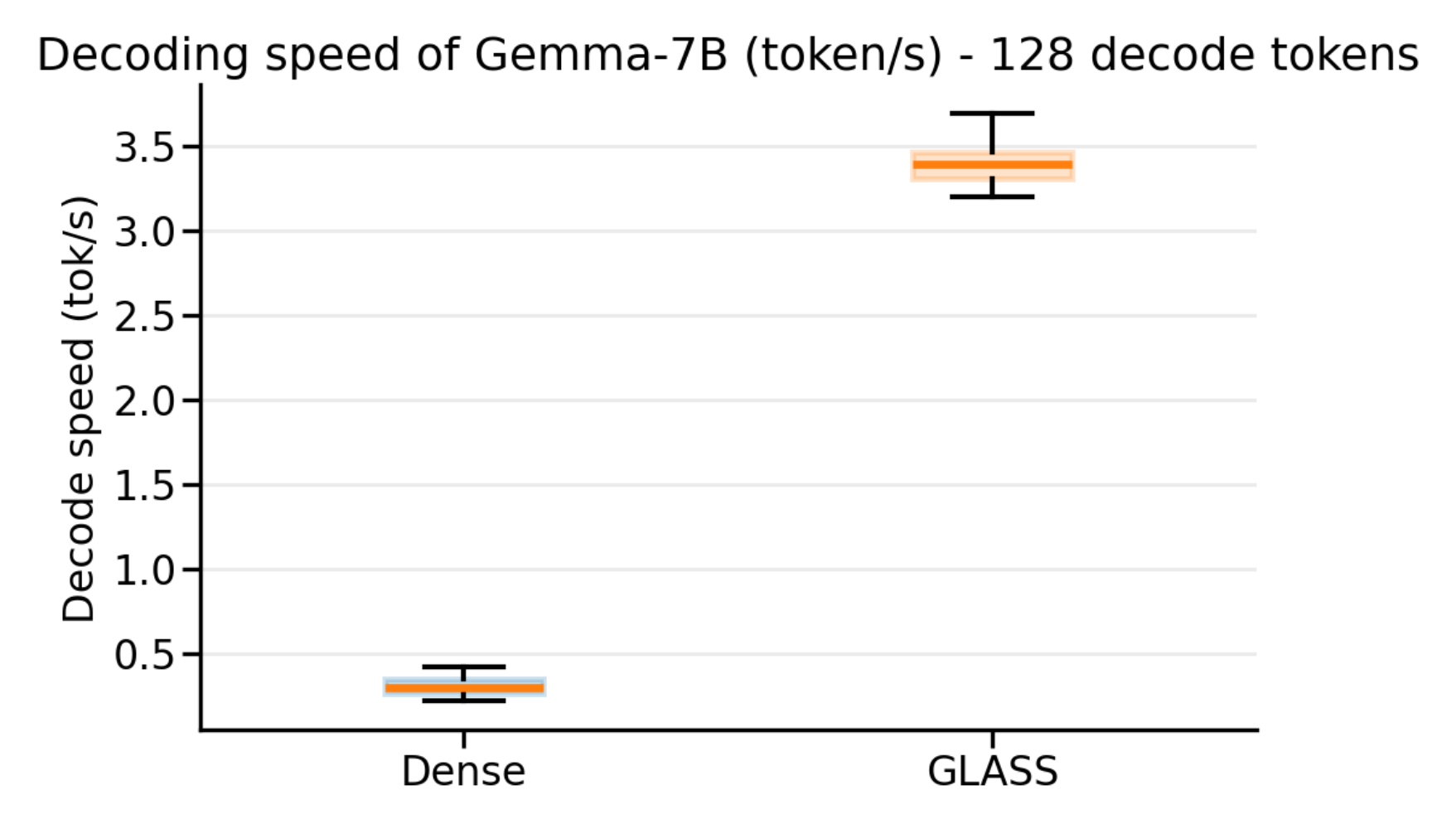}
    \caption{ }
    \label{fig:speedup_128}
\end{subfigure}

\vspace{0.75em}

\begin{subfigure}{0.45\textwidth}
    \centering
    \includegraphics[width=\linewidth]{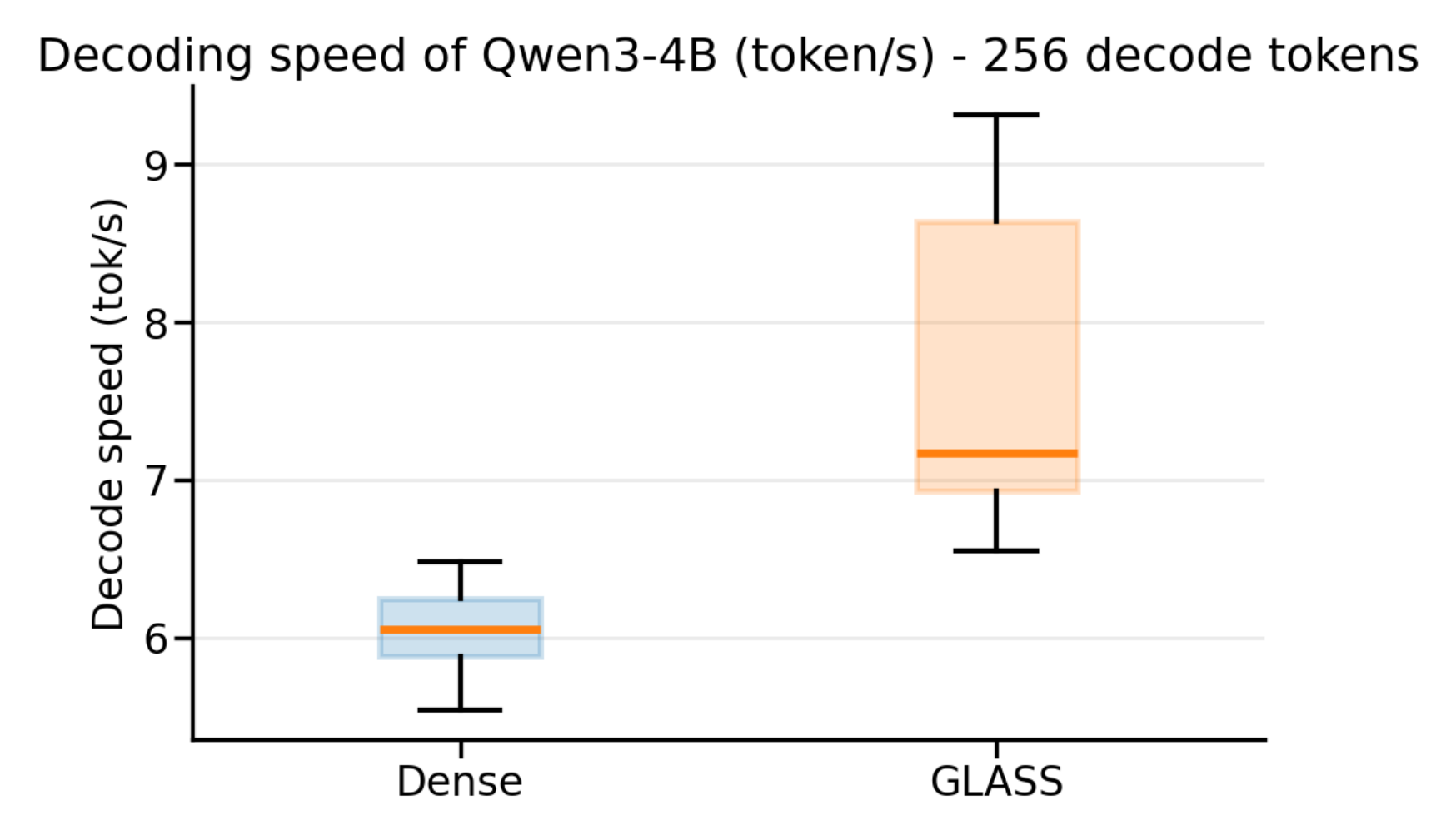}
    \caption{ }
    \label{fig:img2}
\end{subfigure}
\hfill
\begin{subfigure}{0.45\textwidth}
    \centering
    \includegraphics[width=\linewidth]{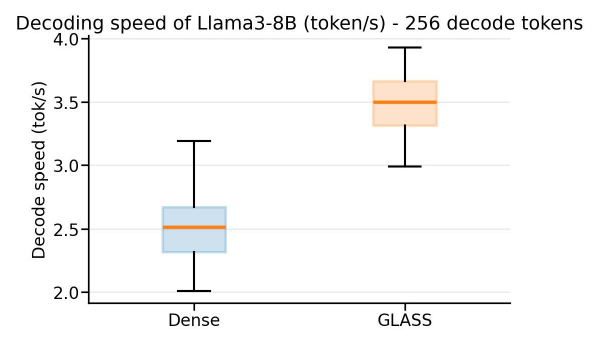}
    \caption{ }
    \label{fig:img4}
\end{subfigure}

\caption{Wall-clock speedup of GLASS compared to the base model with 
  256 tokens generated during the decoding phase for the \hbox{Qwen3 4B} and \hbox{Llama3 8B} models (bottom row), and 64 and 128 generated tokens for the \hbox{Gemma 7B} model (top row).}
  \label{fig:2x2}
\end{figure}

Figs.~\ref{fig:speedup_64}~and~\ref{fig:speedup_128} report the on-device runtime performance of Gemma 7B evaluated on a Samsung Galaxy S25 Ultra with 12\,GB of RAM. Compared to Qwen3 4B and Llama3 8B (Figs.~\ref{fig:img2}~and~\ref{fig:img4}), Gemma 7B has higher memory requirements, particularly during the decoding phase. As a result, the dense version of the model does not fit entirely in device memory, leading to significantly slower execution. For Gemma 7B, we therefore use a prefill length of 64 tokens and decoding lengths of 64 and 128 tokens, as larger settings (e.g., 1024 prefill tokens and 256 decode tokens) exceed the device's memory constraints and cannot be executed reliably.

In contrast, the GLASS variant reduces the effective FFN size by 50\% during decoding, which enables the model to reside fully in RAM. By reducing memory usage, GLASS shows a substantially larger performance gap relative to the dense variant compared to the other two models. Specifically, GLASS achieves an approximately 11$\times$ decoding speedup over the dense Gemma 7B baseline, which is considerably higher than the improvements observed for Qwen3 4B and Llama3 8B. \rD{This case illustrates that on memory-constrained edge hardware, the I/O reduction enabled by structured FFN sparsity can be the dominant source of speedup, beyond the direct compute reduction.}

\end{document}